\documentclass[10pt,twocolumn,letterpaper]{article}

\usepackage[pagenumbers]{cvpr} 

\usepackage[dvipsnames]{xcolor}

\usepackage{url}
\usepackage{graphicx}
\usepackage{algorithm}
\usepackage{algorithmic}
\usepackage{subcaption}
\usepackage{floatflt}
\usepackage{amsfonts}
\usepackage{amssymb}
\usepackage{amsmath}
\usepackage{wrapfig}

\usepackage[export]{adjustbox}

\newcommand{\R}{\mathbb{R}}
\definecolor{cvprblue}{rgb}{0.21,0.49,0.74}
\usepackage[pagebackref,breaklinks,colorlinks,citecolor=cvprblue]{hyperref}

\usepackage[framemethod=TikZ]{mdframed}
\mdfsetup{skipabove=10pt,skipbelow=5pt,roundcorner=4pt}

\definecolor{light-gray}{gray}{0.95}
\usepackage[most]{tcolorbox}
\newtcbox{\mymath}[1][]{%
    nobeforeafter, math upper, tcbox raise base,
    enhanced, colframe=blue!30!black,
    colback=blue!30, boxrule=1pt,
    #1}

\def\ie{\emph{i.e}\onedot} 

\def\etal{\emph{et al}\onedot}

\newcommand{\firstbest}[1]{{\textbf{#1}}}
\newcommand{\secondbest}[1]{{\underline{#1}}}

\DeclareMathOperator{\crossattn}{crossatten}

\DeclareMathOperator{\conv}{conv}
\DeclareMathOperator{\flatten}{flat}
\DeclareMathOperator{\unflatten}{unflat}
\DeclareMathOperator{\relu}{ReLU}

\newcommand{\bv}[1]{\mathbf{#1}}

\def\paperID{4} 
\def\confName{CVPR}
\def\confYear{2024}

\title{Spatio-Temporal Attention and Gaussian Processes for Personalized Video Gaze Estimation}

\author{
Swati Jindal\textsuperscript{1} \hspace{3mm}
Mohit Yadav\textsuperscript{2} \hspace{3mm}
Roberto Manduchi\textsuperscript{1} \\[1.75mm]
\textsuperscript{1}University of California Santa Cruz \quad
\textsuperscript{2} University of Massachusetts Amherst
}

\begin{document}
\maketitle
\begin{abstract}
Gaze is an essential prompt for analyzing human behavior and attention. Recently, there has been an increasing interest in determining gaze direction from facial videos. However, video gaze estimation faces significant challenges, such as understanding the dynamic evolution of gaze in video sequences, dealing with static backgrounds, and adapting to variations in illumination. To address these challenges, we propose a simple and novel deep learning model designed to estimate gaze from videos, incorporating a specialized attention module. 
Our method employs a spatial attention mechanism that tracks spatial dynamics within videos. This technique enables accurate gaze direction prediction through a temporal sequence model, adeptly transforming spatial observations into temporal insights, thereby significantly improving gaze estimation accuracy. 
Additionally, our approach integrates Gaussian processes to include individual-specific traits, facilitating the personalization of our model with just a few labeled samples. 
Experimental results confirm the efficacy of the proposed approach, demonstrating its success in both within-dataset and cross-dataset settings. Specifically, our proposed approach achieves state-of-the-art performance on the Gaze360 dataset, improving by $2.5^\circ$ without personalization. Further, by personalizing the model with just three samples, we achieved an additional improvement of $0.8^\circ$. The code and pre-trained models are available at \url{https://github.com/jswati31/stage}.
\end{abstract}    
\section{Introduction}
\label{sec:intro}

The human gaze is an essential cue for conveying people’s intent, making it promising for real-world applications
such as human-robot interaction~\citep{moon2014meet, palinko2016robot}, AR/VR~\citep{patney2016towards,padmanaban2017optimizing}, and saliency detection~\citep{rudoy2013learning,parks2015augmented}. 
In addition,  gaze plays a vital role in several computer vision tasks, including but not limited to object detection~\citep{vasudevan2018object},
visual attention~\citep{chong2018connecting} and action recognition~\citep{min2021integrating}. 
Despite the primary research emphasis on gaze estimation from images, the potential benefits of understanding the temporal dynamics of eye movements for video gaze estimation have been relatively overlooked. 
Constructing an accurate video-based gaze estimation model requires addressing the unique challenges inherent to videos. These include the evolution of eye movements throughout the video, correlations between gaze directions in successive frames, the predominance of a static background in most pixels, and variations due to individual-specific traits ~\citep{liu2018differential,park2019few,linden2019learning}. 
This work responds to these challenges by aiming to develop an accurate gaze estimation technique for videos using deep networks.

Realizing the potential of spatial and motion cues in videos, 
prior research has utilized residual frames and optical flows for several other vision tasks~\citep{simonyan2014two, feichtenhofer2016convolutional, wang2018temporal}. Specifically, these methods integrate RGB and residual frames as different input streams, requiring larger models with higher inference time and memory requirements~\citep{karpathy2014large, wang2015action, girdhar2017actionvlad}. Similarly, 3D convolutional neural networks (CNNs) can also capture spatiotemporal information from 
videos, but they require many model parameters~\citep{ji20123d, tran2015learning, wang2017two,carreira2017quo, feichtenhofer2019slowfast, li2020spatio}. 
In addition, it is non-trivial to transfer knowledge from pre-trained 3D CNNs to new video tasks, as most pre-trained models rely on large 2D image datasets such as the ImageNet dataset~\citep{deng2009imagenet}.
Despite the critical role of detecting spatial and motion cues in videos, there is a strong need to design efficient attention-based approaches for video-related tasks, including video gaze estimation.

\begin{figure}[t]
    \centering
    \includegraphics[width=\columnwidth]{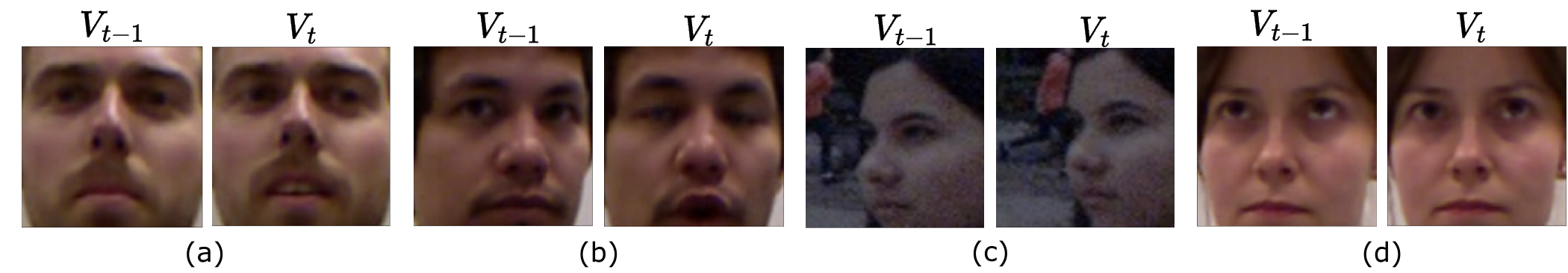}
    \caption{
    The figure illustrates a range of irrelevant factors for video gaze estimation, also referred to as distractors: (a) and (b) depict alterations in facial expression, (c) highlight background movement, and (d) represent a scenario without any distractors.
    These examples show the importance of accurately distinguishing between spatial changes due to eye movements and irrelevant distractors for the video gaze estimation task.
    }
    \label{fig:distractors}     
\end{figure}

In this work, we draw inspiration from the \textit{change captioning} task to develop an approach for video gaze estimation. The change captioning task requires describing the changes between a pair of before and after images, expressed through a natural language sentence~\citep{park2019robust, qiu2021describing, tu2021semantic}. Both change captioning and gaze estimation tasks require differentiating irrelevant distractors, such as background movement and facial expression changes, from the relevant ones. Specifically, change captioning focuses on recognizing object movements, whereas gaze estimation concentrates on detecting eye movements. Similar to prior works~\citep{park2019robust, qiu2021describing}, our approach utilizes a spatial attention mechanism to focus on gaze-relevant information while minimizing the impact of distractors. For example, Figure~\ref{fig:distractors} illustrates various distractors that may obfuscate gaze information in videos.

We introduce \textit{\textbf{S}patio-\textbf{T}emporal \textbf{A}ttention for \textbf{G}aze \textbf{E}stimation (STAGE)}, a deep learning model for video gaze estimation. STAGE utilizes spatial changes in consecutive frames to integrate motion cues via a Spatial Attention Module (SAM) and captures global dynamics with a Temporal Sequence Model (TSM). The SAM module focuses on gaze-relevant information by applying local spatial attention between consecutive frames and effectively suppresses irrelevant distractors. Meanwhile, the TSM considers global dynamic movements across the temporal dimension, enabling enhanced prediction of gaze direction sequences. STAGE adeptly encodes motion information through the attention modules with fewer parameters than existing approaches like 3D CNNs~\citep{ji20123d} or two-branch networks~\citep{karpathy2014large}, thus offering a more feasible solution for real-world applications.

To enhance the accuracy of gaze estimation models, previous studies have suggested personalization to address significant variability in individual-specific traits, such as eye geometry and appearance~\citep{liu2018differential,offsetcalibration, park2019few}. Concretely, this is done by training a person-agnostic gaze model on a large labeled dataset and then fine-tuning it for individual users with a small set of labeled data. Consistent with this approach, we integrate Gaussian processes (GPs)~\citep{rasmussen2004}, known for their effectiveness in low-data scenarios, to personalize the STAGE model for individual users.

We use GPs to learn an additive bias correction and personalize the gaze estimate of the general STAGE model with just a few labeled samples. GPs enable the estimation of personalized 3D gaze directions and provide uncertainty measurements in interval form. These intervals represent a range of possible gaze directions instead of a single vector, making our approach more suitable for practical applications, such as monitoring attention on screens~\cite{zhang2017look, albiz2023guiding}. To evaluate the efficacy of the proposed STAGE model and personalization using GPs, we use three public video gaze datasets: EYEDIAP~\citep{funes2014eyediap}, Gaze360~\citep{kellnhofer2019gaze360} and EVE~\citep{park2020towards}. 

Our primary contributions are as follows:
\begin{itemize}
   \item We introduce STAGE, a novel model for video gaze estimation. STAGE leverages an attention mechanism that is sensitive to spatial changes in sequential frames, effectively extracting gaze-relevant details from videos and facilitating gaze prediction along the temporal axis.
    \item We propose a sample-efficient approach to personalize STAGE, aiming to learn a bias correction model for gaze prediction using pre-trained Gaussian processes~\citep{pretrainedGPs}. 
    \item Our approach either surpasses or matches to the state-of-the-art performance on three publicly available datasets for video gaze estimation. In particular, we obtain state-of-the-art results on the Gaze360 dataset in both cross-data and within-data experimental settings.
\end{itemize}

\section{Related Work}

Traditional methods of gaze estimation use an eye geometry model and exploit regression function to create a mapping from the eye or face images to the gaze vector~\citep{guestrin2006general,hansen2009eye,valenti2011combining, nakazawa2012point,lu2016estimating,kar2017review}. While these methods perform well in controlled settings with consistent subject features, head positions, and lighting, their precision tends to drop in more varied and less controlled environments~\cite{zhang2019evaluation}. 

Recently, with the emergence of deep learning methods, researchers employ CNNs to predict gaze direction from eye or face images directly~\citep{tan2002appearance, zhang2015appearance, krafka2016eye, huang2017tabletgaze, fischer2018rt, cheng2020gaze}. 
Image-based gaze estimation methods primarily use eye images to predict gaze directions~\citep{zhang2015appearance, park2018deep, park2018learning, lian2018multiview}. Additionally, several approaches consider facial features such as head pose and facial appearance for estimating gaze information~\citep{krafka2016eye, zhang2017s, ren2021gaze, gu2021gaze}. Generally, facial information for gaze estimation yields more accurate results than methods relying solely on eye images~\citep{zhang2017s}. Similarly, our work also relies on full-face images for extracting gaze information.

Following the release of video gaze datasets~\cite{funes2014eyediap, kellnhofer2019gaze360}, several temporal gaze estimation models have emerged. These models are designed to predict the gaze direction from a sequence of images. The initial work of \citet{palmero2018recurrent} employs a recurrent CNN that concatenates the static features of each frame and feeds into a recurrent module, which is used to predict the 3D gaze direction of the final frame in the sequence. \citet{kellnhofer2019gaze360} proposed a bidirectional LSTM that utilizes both past and future frames, indirectly incorporating spatial information. 
\citet{wang2019neuro} released a dataset that captures eye images and ground-truth gaze positions on a screen while subjects engage in activities like browsing websites or watching videos. They proposed a dynamic gaze transition network to detect the transitions of eye movements over time and refine static gaze predictions using the dynamics learned from these transitions. Similarly, \citet{park2020towards} collected a large video gaze dataset and proposed a recurrent model to refine Point of Gaze (PoG) estimates for video input. Our work aims to develop a video gaze estimation method by capturing the nuanced spatial and temporal dynamics.

As stated earlier, the performance of gaze estimators can be notably influenced by individual-specific traits, particularly when adapting these models to new subjects~\citep{guestrin2006general}. However, in practical scenarios, there are typically only a few labeled samples available per subject and are insufficient for fine-tuning contemporary deep learning models, which tend to be over-parameterized~\citep{park2019few}. Previously, ~\citet{liu2018differential} utilized a Siamese network to estimate gaze differences, employing a small number of calibration samples for personalization. Similarly,~\citet{park2019few} employed meta-learning techniques to achieve few-shot personalization, leveraging learned gaze embeddings. \citet{offsetcalibration} introduced a method to model person-specific biases during the training phase, enabling personalization during testing with just a few samples. Our personalization approach is motivated by the efficacy of Gaussian processes in scenarios with limited data~\citep{rasmussen2004}.
Unlike \citet{offsetcalibration}, our personalization approach outputs a different bias for each video frame and is designed to be compatible with any existing gaze estimation technique without necessitating alterations to the training objective.

\section{Proposed Method}
\label{sec:proposed_section}


\begin{figure}[t]
    \centering
    \includegraphics[trim=1.9cm .0cm 1.8cm .0cm,clip=true,scale=0.62]{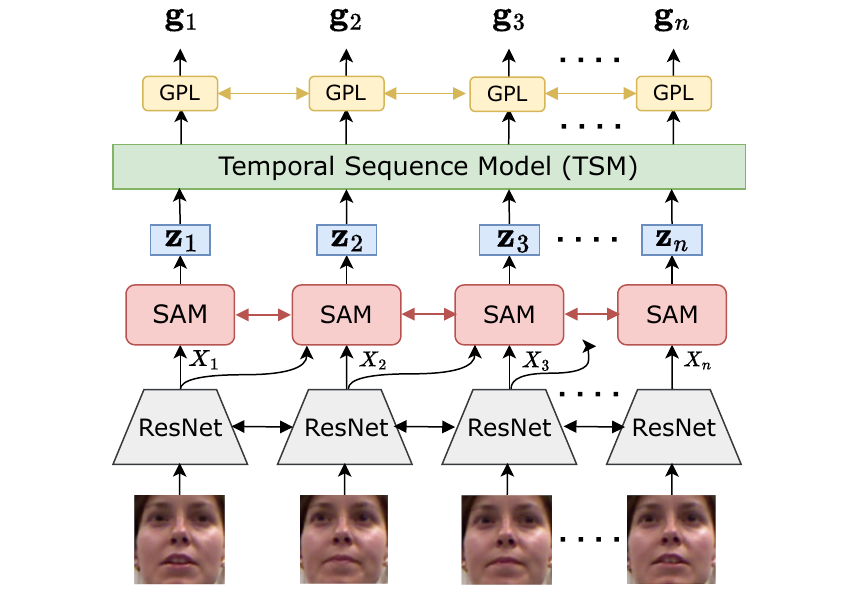}
    \caption{\textit{A schematic overview of the proposed (person-agnostic) STAGE model.} The proposed model has three modules: spatial attention module (SAM),  temporal sequence model (TSM), and gaze prediction layer (GPL). The SAM is designed to extract information relevant to the gaze by concentrating on the spatial differences between consecutive frames. In the figure, ${X}_i$ represents features from ResNet, $\bv{z}_i$ denotes the motion-informed output of the SAM, and $\bv{g}_i$ corresponds to the predicted gaze direction.}
    \label{fig:arch}     
\end{figure}

The main goal of video gaze estimation is to learn a deep network $f$ defined as  $f: V \mapsto G$ that maps a sequence of video frames $V \in \mathbb{R}^{n \times h_0 \times w_0 \times 3}$ to a sequence of gaze directions $G \in \mathbb{R}^{n \times 2}$, where $n$ is the number of frames and $h_0$ and $w_0$ are height and width of each frame, respectively. The output gaze sequence $G$ possesses pitch and yaw angles, which correspond to each frame in $V$.

The proposed STAGE model employs three modules for setting up the deep network $f$.
Firstly, a ResNet-based CNN model receives the input video and extracts feature maps for all the frames. 
Then, in the following module of the STAGE model, we process feature maps using a \textit{Spatial Attention Module} (SAM) to focus on the spatial motion information between consecutive frames followed by a \textit{Temporal Sequence Model} (TSM)  to learn temporal dynamics using past frame embeddings. Next, the gaze prediction layer (GPL) maps the features from the output of the TSM block to a sequence of gaze directions defined in terms of yaw and pitch angles. Figure~\ref{fig:arch} shows the schematic of STAGE.

\begin{figure*}[t]
    \centering
    \includegraphics[trim=1.8cm 0.2cm 1.2cm 0.0cm,clip=false, scale=0.58]{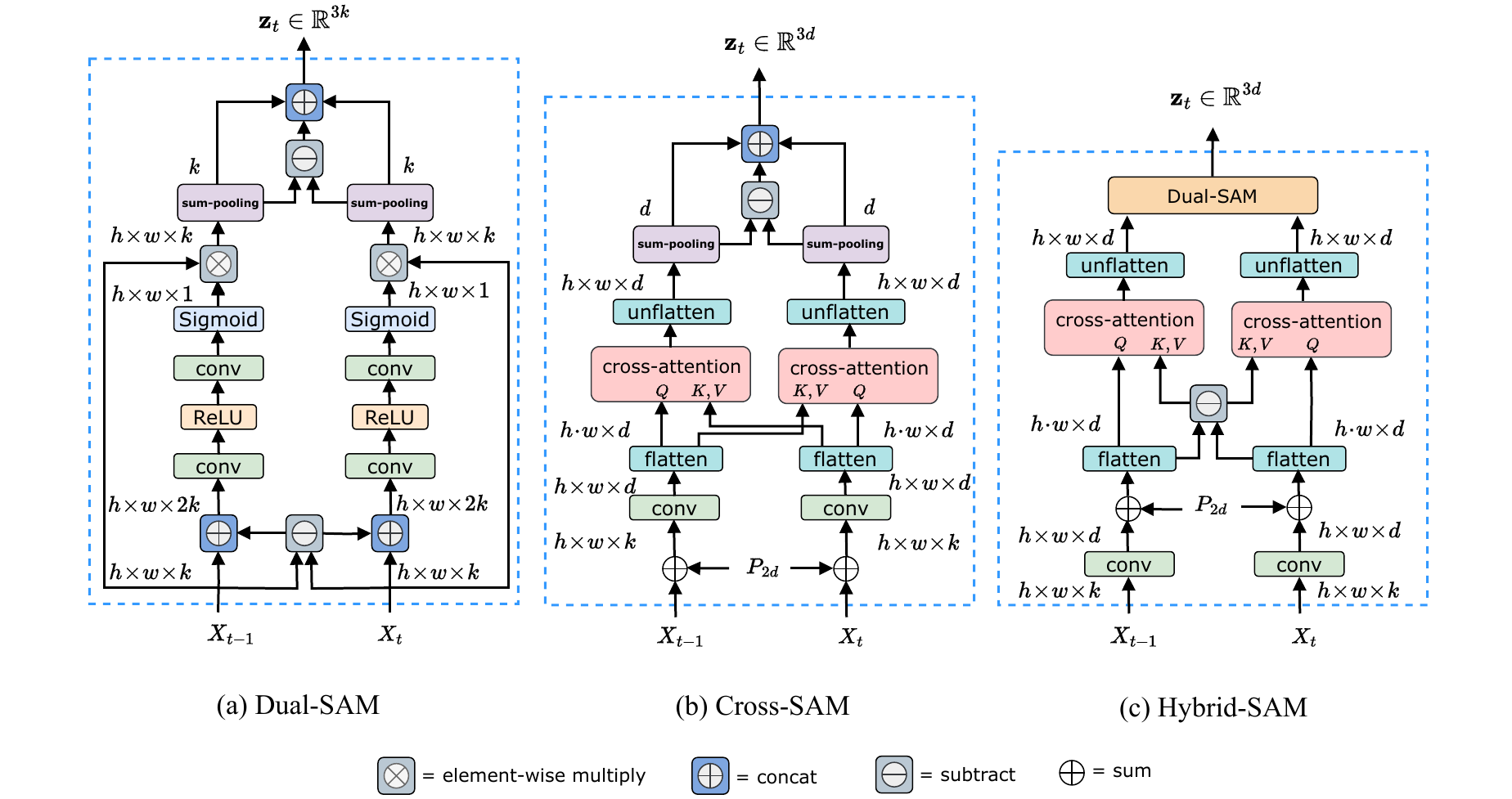}
    \caption{\textbf{Block diagram of SAM variants.} The input in each SAM variant is a pair of the consecutive frame features $X_{t-1}$ and $X_t$, and the output is a 1-D feature vector encoding both RGB and motion information. 
    ${P}_{2d}$ are 2D positional embeddings with height and width that are the same as the input feature map. 
    The \textit{cross-attention} block in Cross-SAM and Hybrid-SAM is a standard transformer operation. 
    The \textit{sum-pooling} block applies feature pooling by summing them over height and width dimensions. More details are in Section~\ref{ssec:spatial_attnetion_model}.
    }  
    \label{fig:allsams}
\end{figure*}

\subsection{Spatial Attention Module (SAM)}
\label{ssec:spatial_attnetion_model}

Recall that SAM is aimed to distinguish gaze-relevant motion by analyzing differences between consecutive frames, focusing on crucial cues like eye or head movements for gaze estimation while filtering out irrelevant distractions like facial expressions or background movements. It aims to prioritize relevant video changes, particularly eye movements, and disregard non-essential ones.

First, we convert each frame 
in the video sequence $V$ to features 
$X = \left[X_1, X_2, \ldots, X_n \right] \in \mathbb{R}^{n\times h \times w \times k}$, using the ResNet-based CNN model, where $w$, $h$, and $k$ are the width, 
height, and the number of channels of the feature maps extracted by ResNet. The next step is to pass each consecutive feature pair $(X_{t-1}, X_{t})$ through a shared SAM. 
Concretely, the SAM module aggregates information from  RGB features of $X_{t-1}$ and $X_{t}$, and the feature differences $(X_{t} - X_{t-1})$ through a fusion strategy. 
Figure~\ref{fig:allsams} provides an overview of all three SAM variants considered in this work. All SAM variants are optimized during model training and outputs $\mathbf{z}_t$, a feature representation with spatial motion information for the $t^{\text{th}}$ frame of the video. 

\textbf{Dual-SAM} predicts separate spatial attention maps for both current $X_{t}$ and past $X_{t-1}$ frame. It compares the spatial attention maps of the current and past frames and 
identifies the
region that is most relevant to the observed motion changes. If the spatial attention maps are very similar, SAM infers that there is no substantial change between consecutive frames and encodes these minimal differences in the output vector $\mathbf{z}_t \in \mathbb{R}^{3k}$. Conversely, if there is a difference, SAM incorporates this change into the output vector $\mathbf{z}_t$. This SAM variant is inspired by \citet{park2019robust} in the change captioning task and is shown in Figure-\ref{fig:allsams}a.

\textbf{Cross-SAM}, unlike {Dual-SAM}, utilizes cross-attention from transformer models~\cite{vaswani2017attention} to encapsulate dense correlation between each pair of image patches in the past and current frames. This allows {Cross-SAM} to identify multiple changes between two frames, as opposed to {Dual-SAM}, which can only capture a single change. Practically,  detecting multiple changes and subsequently filtering out irrelevant distractors is more useful for video gaze estimation tasks. 
Similar to the {Dual-SAM}, this variant utilizes both RGB and transformed motion signals at the output. {Cross-SAM} is motivated by \cite{qiu2021describing} and is shown in Figure-\ref{fig:allsams}b.

\textbf{Hybrid-SAM} combines the strengths of both {Dual-SAM} and {Cross-SAM} variants. {Dual-SAM} focuses on one local change, while {Cross-SAM} focuses on global context and captures multiple changes. Similar to Cross-SAM, Hybrid-SAM encapsulates multiple changes by applying a cross-attention mechanism using global context through position embeddings. However, unlike the {Cross-SAM} variant, it uses the difference between current and past frames as a key and value, emphasizing regions with the most significant motion differences. The Dual-SAM is utilized as a pooling operator to selectively focus on the most relevant changes, like eye or head movements, which are crucial for the task of gaze estimation.

The 
{Hybrid-SAM} is given in the Algorithm~\ref{alg:proposed_sam}, and {Dual-SAM} and {Cross-SAM} are deferred to supplementary material. The input is features of the past frame $X_{t-1}$ 
the and current frame $X_{t}$, respectively.
Both input features are projected to the higher-dimensional feature maps using the convolution operation, and 2-D position embeddings $P_{2d} \in \mathbb{R}^{h \times w}$ are added (Line 1). Line 2 computes difference features $X_{\text{diff}}$ for the video's $t^{th}$ frame and cross-attention is applied in Line 3. Lines 5-8 correspond to the same operations as in {Dual-SAM}. 
In Algorithm~\ref{alg:proposed_sam}, $\sigma$ denotes the sigmoid function, $\mathbf{1}_{h, w}$ is a one-hot vector spanning the spatial dimensions, and $\odot$ is an element-wise dot product.

\begin{algorithm}[t] 
\caption{Hybrid-Spatial Attention Module}\label{alg:proposed_sam}
\begin{algorithmic}[1]
\REQUIRE  $X_{t-1}, X_{t} \hfill \in \mathbb{R}^{h \times w \times k}$
\ENSURE  $\mathbf{z}_{t} \hfill \in \mathbb{R}^{3 \cdot d}$
\STATE $X_{t-1} \quad =  \flatten( \conv(X_{t-1}) + \mathbf{1}_{h, w} \odot P_{2d})$ \newline 
 $X_{t} \quad \quad =  \flatten(\conv(X_{t})  + \mathbf{1}_{h, w} \odot  {P}_{2d})\hfill \in \mathbb{R}^{h \cdot w \times d}$ 
\STATE $X_{\text{diff}} \, \quad= X_{t}  - X_{t-1} $
\STATE $X_{t-1} \quad = \crossattn(X_{t-1}, X_{\text{diff}}, X_{\text{diff}})$ \newline 
$X_{t} \quad \quad = \crossattn(X_{t}, X_{\text{diff}}, X_{\text{diff}})  \hfill\in \mathbb{R}^{ h \cdot w \times d}$
\STATE $X_{t-1} \quad = \unflatten(X_{t-1}, h \times w) $ \newline 
$X_{t} \quad\quad = \unflatten(X_{t}, h \times w)\hfill\in \mathbb{R}^{ h \times w \times d}$
\STATE $X'_{t-1} \quad  = [X_{t-1}; X_{t}-X_{t-1}]$ \newline 
$X'_{t} \quad \quad = [X_{t}; X_{t} - X_{t-1}]  \hfill \in \mathbb{R}^{h \times w \times 2 \cdot d}$
\STATE $A_{t-1} \quad = \sigma (\conv(\relu(\conv(X'_{t-1}))))$ \newline 
${A}_{t} \quad \quad = \sigma (\conv(\relu(\conv({X}'_{t})))) \hfill \in \mathbb{R}^{h \times w \times 1} $ 
\STATE $\mathbf{v}_{t-1} \quad = \sum_{h, w} {A}_{t-1} \odot {X}_{t-1}$ \newline 
$\mathbf{v}_{t} \quad \quad = \sum_{h, w} {A}_{t} \odot {X}_{t} \hfill \in \mathbb{R}^{d}$
\STATE $\mathbf{z}_{t} \quad \quad = [\mathbf{v}_{t-1}; \mathbf{v}_{t} - \mathbf{v}_{t-1}; \mathbf{v}_{t}]  \hfill \in  \mathbb{R}^{3 \cdot d}$
\RETURN $\mathbf{z}_{t}$
\end{algorithmic}
\end{algorithm}

\subsection{Temporal Sequence Model (TSM)}
\label{ssec:temporal_sequence_model}
The temporal sequence model subsumes spatially enhanced representations $\mathbf{z}_{t}$ produced by the SAM module and is intended to capture the temporal dynamics of the eye movements in the video. In particular, we consider two variants for TSM: recurrent neural networks (RNN)~\citep{sutskever2014sequence}, and transformer network~\citep{vaswani2017attention}. The RNN consists of unidirectional LSTM layers~\citep{hochreiter1997long}, and the transformer variant is a causal transformer decoder, which is prevalent in generative language modeling, such as the GPT-2 model~\citep{radford2019language}. We provide more details in supplementary materials.

\subsection{Gaze Prediction Layer and Training Objective}
\label{ssec:gaze_prediction_layer}

The gaze prediction layer is shared across all timestamps and uses an MLP to predict the gaze direction from the frame embeddings generated by the TSM module. 
For $i^{\text{th}}$ sample and $t^{\text{th}}$ frame, let $\{\mathbf{g}_t^{i}\}$ and $\{\mathbf{\hat{g}}_t^{i}\}$ denote the sequences of true and predicted gaze directions, respectively. Similarly, $\{\mathbf{p}_t^{i}\}$ and $\{\mathbf{\hat{p}}_t^{i}\}$ represent the sequences of true and predicted 2D Point-of-Gaze (PoG).
We use the following objective function for training STAGE model parameters (similar to \citet{park2020towards}):
\begin{equation}
\resizebox{\columnwidth}{!}{%
\label{loss:objective}
    $\mathcal{L}_{final} = \dfrac{1}{b \cdot n} \sum_{i=1}^{b}\sum_{t=0}^{n-1} \dfrac{180}{\pi} \arccos \Bigg( \dfrac{{\mathbf{g}_t^{i}}^{T} \mathbf{\hat{g}}_t^{i}}{|\mathbf{g}_t^{i}| \cdot |\mathbf{\hat{g}}_t^{i}|}\Bigg) + \lambda \cdot ||\mathbf{p}^i_{t} - \hat{\mathbf{p}}^i_{t}||$
}
\end{equation}
Here,  
$\lambda$ controls the trade-off between 3D gaze angular error and 2D PoG mean absolute error.



\subsection{Personalizing STAGE using Gaussian Processes}
\label{ssec:personalization_using_gaussian_processes}

As previously mentioned, we propose person-specific Gaussian processes for modeling bias correction terms for each user, which operates on top of the proposed (person-agnostic) STAGE model. 
Specifically, if $f: V \mapsto G$ is the STAGE model, 
then the final prediction for person $p$ is $\hat{\bv{f}}_p(V) =  \bv{f}(V) + \bv{r}_p(V)$, where $\bv{r}_p$ is GP-based bias correction model for the person $p$, \ie, it predicts the residual in addition to the model-agnostic prediction. The GP $\bv{r}_p$ 
models the components of gaze direction (\ie, yaw and pitch) independently at the frame level, using two one-dimensional independent GPs.
Concretely, $\bv{r}_p(V) = [(r_{p, \theta}(V_1), r_{p, \phi}(V_1)), (r_{p, \theta}(V_2), r_{p, \phi}(V_2)), \cdots, (r_{p, \theta}(V_n)$, \newline $r_{p, \phi}(V_n))]$, where $r_{p, \theta}$ and $r_{p, \phi}$ are the one-dimensional GP predictions for pitch and yaw components, respectively.

For GP hyper-parameter tuning and inference, we collect a set of training frames $\mathcal{D} = \{\bv h_i, y_i\}_{i=1}^{\ell}$ 
that are available for person $p$, where $\bv h_i \in \R^d$ are the flattened ResNet output features from the  STAGE model, and $y_i$ is either pitch or yaw of residual gaze angle, \ie, $ \bv{{g}}_i - \bv{\hat{g}}_i$, where, $\bv{{g}}_i$  and $\bv{\hat{g}}_i$ are true gaze direction and 
STAGE's predicted direction, respectively. To represent the dataset $\mathcal{D}$ in matrix format, we let $\bv{y} \in \R^{\ell}$ be the vector of residual angles, where the $i^{th}$ entry equal to $y_i$, and $H \in \R^{\ell \times d}$ have its $i^{th}$ row equal to the ResNet features $\bv h_i$. For brevity, we omit the person index $p$ from henceforth discussion on GPs. 

A Gaussian process associated with kernel (covariance) function $k(\mathbf{h}, \mathbf{h}^{\prime}): \R^d \times \R^d \rightarrow \R$ is a distribution over functions that maps features to residual angles such that, for any $\bv h_1,\ldots, \bv h_\ell \in \R^d$:
\begin{align*}
\label{eq:gp_prior}
\bv{r} = [r(\mathbf{h}_1),..., r(\mathbf{h}_\ell)] \sim \mathcal{N}(\bv{\mu}_0, K_H),
\end{align*}
where $K_H = [k(\bv{h}_i, \bv{h}_j)]_{i,j=1}^{\ell} \in \mathbb{R}^{\ell \times \ell}$ is the kernel (covariance) matrix on the data points $H$,
and $r$ has a constant mean function with its value set to  $\bv{\mu}_0$.
The observed residual angle $y_i$ is modeled as the i.i.d. Gaussian noise, \ie, $y_i \sim \mathcal{N}(r(\bv h_i), \sigma^2 I)$. 
In particular, we use the (squared-exponential) automatic-relevance-determination (ARD) kernel, given as $k(\bv{h},\bv{h}') = \tau \cdot e^{-  \sum_{s=1}^{d}
\frac{ \left(\bv{h}(s) - \bv{h}^{'}(s)\right)^2}{\theta(s)^2}}$, where 
$\tau$ and $\theta \in \mathbb{R}^d$ are kernel hyper-parameters. The ARD kernel's per-dimension scaling, being more expressive than the RBF kernel's use of a single length-scale, often leads to superior practical performance~\citep{ard_kernel}. Intuitively, this flexibility allows the model to adapt to varying feature relevance and noise levels, potentially leading to improved accuracy and generalization~\citep{ard_kernel2}. Upon conditioning the GP model on the collected training dataset, the predictive posterior mean and covariance functions are as follows:
\begin{align*}
& \textbf{mean: } \mu_{r|\mathcal{D}}(\mathbf{h}) = \bv{k}_\bv{h}^T ({K}_H +\sigma^2 I)^{-1} \mathbf{y}\\ 
& \textbf{variance: }  \sigma_{r|\mathcal{D}}(\mathbf{h}) = k(\mathbf{h}, \mathbf{h})
- \bv{k}_{\bv{h}}^T ({K}_H +\sigma^2 I)^{-1}  \bv{k}_{\mathbf{h}}\nonumber
\end{align*}
where the vector $\bv{k}_{\bv h} \in \R^{\ell}$ has $i^{th}$ entry $ k(\bv{h},\bv{h}_i)$, \ie, kernel value between any feature vector $\bv h$ and $i^{th}$ data point. The posterior mean function predicts the residual gaze angles and is utilized for correction. The posterior covariance function determines the uncertainty in this prediction, as illustrated in Figure~\ref{fig:visual}.



\paragraph{Optimizing GP hyper-parameters using a  few labeled samples.} 
GPs are non-parametric models and thus do not require tuning many parameters \citep{rasmussen2004}. However, they still necessitate optimizing hyperparameters, which in our case are ${\mu}_0$, $\sigma$, $\tau$, and $\theta$, totaling $d+3$ hyperparameters as $|\theta| = d$. The ARD kernel adds flexibility to the GP model but also increases the number of hyperparameters to be tuned. Specifically, since $d=16384$ when using features from the ResNet model, directly tuning hyperparameters using the log-likelihood of data $\mathcal{D}$ is prone to overfitting, particularly when as few as three samples are present in $\mathcal{D}$. To overcome this challenge, we propose the application of pre-trained GPs, similar to the concurrent work \cite{pretrainedGPs}. Pre-trained GPs entail the initial optimization of hyperparameters on data used for training the STAGE model, coupled with the implementation of early stopping to maximize the log-likelihood of dataset $\mathcal{D}$ for each individual. This methodology grants GPs flexibility with expressive ARD kernel and ensures a robust starting point due to pre-training. 

\section{Experiments and Results}


\paragraph{Datasets.}

{EVE}~\citep{park2020towards} is a large video gaze dataset comprising over 12M frames collected from 54 participants in a controlled indoor setting with four synchronized and calibrated camera views. Following the splits in \cite{park2020towards}, there are 40 subjects in training and 6 in the validation set. We discard the test subjects due to the unavailability of labels and evaluate our models on the validation set. 
{Gaze360}~\citep{kellnhofer2019gaze360} is a large-scale, physically unconstrained gaze dataset collected from 238 subjects in indoor and outdoor settings. The dataset includes a wide range of head poses, with 129K training, 17K validation, and 26K test images. We evaluate our models on all three subsets of the dataset: the full Gaze360 dataset, the front $180^\circ$ subset, and the front $20^\circ$ subset, as done in \cite{kellnhofer2019gaze360}. 
{EyeDiap}~\citep{funes2014eyediap} consists of 94 videos totaling 237 minutes, collected from 16 subjects in a laboratory environment. 
It includes videos for both screen and floating targets, and we select VGA videos of screen targets. 

\paragraph{Implementation Details.} The input video sequence $V$ consists of 30 frames containing a full-face image of $128\times128$ pixels. We use ResNet-18~\citep{shafiq2022deep} initialized with GazeCLR~\citep{jindal2022contrastive} weights shared between all timestamps to extract visual features from the image sequence. The third convolutional layer block of ResNet-18 outputs features with a dimension of $256\times8\times8$. We pass these features through the SAM module, followed by TSM and gaze prediction layers. 
We train STAGE end-to-end for $50$K iterations using the SGD optimizer with an initial learning rate of 0.016 and momentum of 0.9. The learning rate is decayed using cosine annealing~\citep{loshchilov2016sgdr}, and the batch size is set to 16. We discuss the implementation in more detail in the supplementary.

\subsection{Evaluating the STAGE Model}


\paragraph{Baselines.}
We benchmarked our framework against EyeNet~\citep{park2020towards}, which consists of ResNet-18 and RNN layers and uses both eye image patches as input. We adopted EyeNet to our setting and trained it on full-face images using  $\mathcal{L}_{final}$ with $\lambda=0.001$. 
We also train another variant of EyeNet by replacing the RNN module with a TSM similar to that used in our framework. For a fair comparison, we also implement EyeNet with our version of ResNet-18 initialized with GazeCLR~\citep{jindal2022contrastive} weights and call it \textit{EyeNet (GazeCLR)}. Further, we adapt \cite{chang2021mau} for gaze estimation, which introduces motion-aware-unit (MAU) for the video-prediction task. We also compare with a simple baseline by removing the SAM modules and concatenating $X_t$ and $X_{\text{diff}} = (X_t - X_{t-1})$ before passing through TSM, termed \textit{Concat-Residual}. Finally, we compare the three variants of SAM combined with two variants of TSM for cross-dataset and within-dataset experiments. For the sake of completion, we also evaluate the Hybrid-SAM method without the Dual-SAM module at the output, named as {Hybrid-$\text{SAM}^{\dagger}$}.

\begin{figure}[t]
    \centering
    \includegraphics[width=\columnwidth]{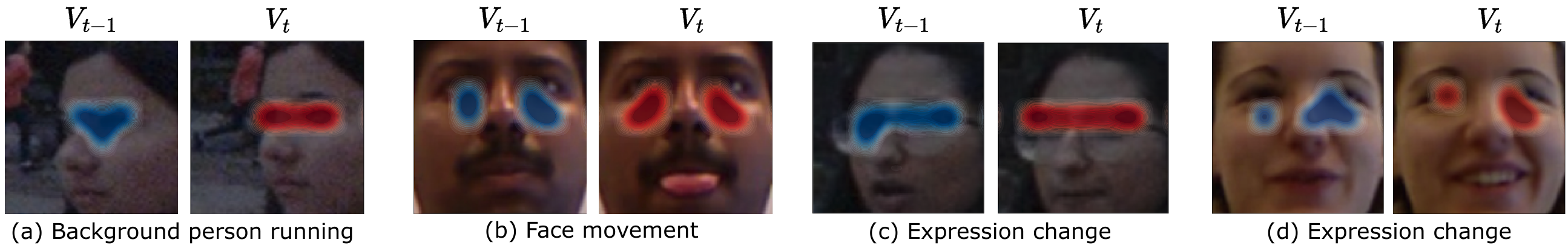}
    \caption{Illustration of attention maps $A_{t-1}$ and $A_{t}$, generated by the Hybrid-SAM, superimposed on sequential video frames $V_{t-1}$ and $V_{t}$. The SAM module proficiently highlights the ocular area, key for analyzing eye movements, while simultaneously diminishing irrelevant distractions such as background motion (a), tongue movement (b), and changes in emotional expressions (c and d).
    }
    \label{fig:qual-visual}     
\end{figure}

\subsubsection{Qualitative Evaluation}
\label{subsec:quality-exp}

We conducted a qualitative analysis primarily centered on assessing the Hybrid-SAM ability to distinguish between gaze-irrelevant distractors and gaze-relevant eye movements, which is crucial for video gaze estimation, as stated earlier. Specifically, we examined attention maps $A_{t-1}$ and $A_{t}$, strategically overlaid on sequential video frames $V_{t-1}$ and $V_{t}$, as depicted in Figure~\ref{fig:qual-visual}. We analyzed several frames showcasing scenarios from background activities to facial movements, all concurrent with dominant eye movements. 

In Figure~\ref{fig:qual-visual}(a), the network adeptly focuses on eye movements in frame $V_t$ (red pixels) and prior frame changes (blue pixels) despite significant background pixel shifts from a walking person. This underscores the effectiveness of spatial attention in filtering out irrelevant distractors to accurately identify subtle eye movements and gaze direction. As a result, it eases the process of temporal modeling in video gaze estimation. Additionally, as illustrated in Figure~\ref{fig:qual-visual}(b), although tongue movement presents a potential distraction, it is efficiently disregarded. Moreover, changes in facial expressions, depicted in Figure~\ref{fig:qual-visual}(c, d), are effectively overlooked by the Hybrid-SAM. These qualitative findings affirm that the spatio-temporal attention strategy adeptly minimizes significant distractions, particularly in the eye region, which is essential for accurately tracking eye movements in video gaze estimation tasks.

\begin{table}[t]
\centering
\begin{tabular}{lccc} 
\hline
\textbf{Method} & \textbf{Full} & $\mathbf{180^\circ}$ & \textbf{Front} $\mathbf{20^\circ}$\\
\hline
%
%
%

EyeNet \citep{park2020towards}(GazeCLR) & 12.53 & 12.08 & 9.45\\
EyeNet + Tx & 13.00 & 12.55 & 9.73 \\
\hline
Concat-Residual + LSTM & 10.35 & 10.16 & 7.45\\
Concat-Residual + Tx & 12.22 & 11.78 & 9.09\\
\hline
Dual-SAM + LSTM & 10.12 & 9.92 & \secondbest{7.08}  \\
Dual-SAM + Tx & 10.13 & 9.93 & 7.23\\
\hline
Cross-SAM + LSTM & 12.00 & 11.59 & 9.51\\
Cross-SAM + Tx & 10.12 &  9.91 & 7.34\\

\hline
Hybrid-$\text{SAM}^{\dagger}$ + LSTM  & 12.69 & 12.26 & 9.66 \\

Hybrid-$\text{SAM}^{\dagger}$ + Tx  & 12.33 & 11.90 & 9.53 \\


Hybrid-SAM + LSTM & \firstbest{10.05} & \firstbest{9.84} & \firstbest{6.92} \\

Hybrid-SAM + Tx & \secondbest{10.10} &  \secondbest{9.90} & 7.33\\
\hline


 
\end{tabular}
\caption{\textbf{Within-dataset Evaluation.} 
Comparison of mean angular errors (in degrees) between the proposed STAGE model, SAM and TSM variants, and other baseline approaches. Tx is the transformer TSM model.
The \textbf{first} and \underline{second} best results are bold-ed and underlined, respectively.
}
\label{table:withindata-eval}
\end{table}

\subsubsection{Within-dataset Evaluation}
\label{subsec:within-exp}
Table~\ref{table:withindata-eval} shows results for within-dataset evaluation, where we train and evaluate our model on the same domain dataset. We train our framework on the training subset of Gaze360 with $\lambda=0$ and evaluate it over three test subsets as done in~\cite{kellnhofer2019gaze360}. Our model demonstrates superior performance compared to the baselines, including `Concat-Residual,' across all three subsets. Specifically, it achieves absolute improvements of $2.5^\circ$, $2.2^\circ$ and $2.5^\circ$ on full Gaze360, front $180^\circ$ and front $20^\circ$ subsets, respectively. Furthermore, it is noteworthy that Hybrid-SAM performs better in comparison to Hybrid-$\text{SAM}^{\dagger}$, illustrating the advantage of incorporating Dual-SAM as a pooling operator.


\subsubsection{Cross-dataset Evaluation} 
\label{subsec:cross-exp}
We performed a cross-dataset evaluation, where the model was trained on the EVE dataset and evaluated on two different datasets, EyeDiap and Gaze360. Table~\ref{table:crossdata-eval} shows the comparison of baselines and our proposed method. We observed a significant improvement in both datasets even with a simple concatenation of $X_t$ and $X_{\text{diff}}$, \ie, {Concat-Residual} approach outperforms EyeNet variants and MAU approach, which demonstrates that residual frames are an effective cue for video-gaze estimation. The Dual-SAM and Cross-SAM show improvements over {Concat-Residual} approach, indicating that the adapted methods are more accurate than naively using residual frames. 
Notably, the Hybrid-SAM improves over
baselines by $1.2^{\circ}$ in absolute and $14.28\%$ in relative on the EyeDiap dataset. It also outperformed Dual-SAM and Cross-SAM on all three evaluation sets. 
The last two columns of Table~\ref{table:crossdata-eval} show results on the full and front $180^\circ$ Gaze360 subsets. The Hybrid-SAM improved up to $3.6^\circ$ on both subsets, further emphasizing the effectiveness of SAM. It is also worth noting that the performance improvements for SAM
hold for both LSTM and transformer-based TSM in both Tables~\ref{table:withindata-eval} and \ref{table:crossdata-eval}. This shows that SAM is helpful irrespective of the choice of TSM model.

\begin{table}[t]
\centering
\begin{tabular}{lccc} 
\hline
\textbf{Method} & \textbf{EyeDiap} & \textbf{Full} & $\mathbf{180^\circ}$ \\ 
\hline
MAU & 21.30  & 34.18 & 33.57  \\
EyeNet~\citep{park2020towards} & 16.07 & 31.37 &  30.77 \\ 

EyeNet (GazeCLR) & 7.74 & 26.57 & 25.95 \\ 

EyeNet + Tx & 8.40 & 26.25 & 25.64 \\ 
\hline
Concat-Residual+ LSTM & 7.12 &  24.12 &  23.52  \\ 
Concat-Residual+ Tx & 7.27 & 24.26 &  23.64  \\ 
\hline
Dual-SAM + LSTM & 7.04 & 24.18 & 23.58 \\ 
Dual-SAM + Tx & 6.77 & 23.99 & 23.38 \\ 

\hline

Cross-SAM + LSTM  & 8.42 & 23.19 & 22.61 \\ 
Cross-SAM + Tx & 8.75 & \firstbest{22.57} & \firstbest{22.01} \\ 

\hline

Hybrid-$\text{SAM}^{\dagger}$ + LSTM & 8.48 & 23.31 & 22.72 \\ 

Hybrid-$\text{SAM}^{\dagger}$ + Tx  & 7.79 & \secondbest{22.66} & \secondbest{22.09} \\ 


Hybrid-SAM + LSTM & \secondbest{6.70} & 23.73 & 23.13 \\ 

Hybrid-SAM + Tx &  \firstbest{6.54} & 23.77 & 23.17  \\ 




\hline  
\end{tabular}
\caption{\textbf{Cross-dataset Evaluation.} 
Comparison of mean angular gaze error (in degrees) between the proposed STAGE model, SAM and TSM variants, and other baseline approaches. Tx is the transformer TSM model. 
For each column, the \textbf{first} best result is bold-ed, and \underline{second} best result is underlined.}
\label{table:crossdata-eval}
\end{table}

\begin{table}[t]
    \centering
    \begin{tabular}{lccc} 
    \hline
    \textbf{Method} & \textbf{Full} & $\mathbf{180^\circ}$ & \textbf{Front} $\mathbf{20^\circ}$\\
    \hline
    Gaze360~\citep{kellnhofer2019gaze360} & 13.50 & 11.40 & 11.10 \\
    
    MSA+Seq~\citep{Mishra2020360DegreeGE} & 12.50 & 10.70 & - \\
    
    \hline
    SwAT~\citep{farkhondeh2022towards} &  11.60 & - & - \\
    
    L2CS-Net~\citep{abdelrahman2022l2cs} & - & 10.41 & 9.02 \\
    GazeTR-Pure~\citep{cheng2022gaze} &  - & 13.58 & - \\
    GazeTR-Hybrid~\citep{cheng2022gaze} &  - & 10.62 & - \\
    
    \hline
    Hybrid-SAM + LSTM & \firstbest{10.05} & \firstbest{9.84} & \firstbest{6.92} \\
    Hybrid-SAM + Tx & \secondbest{10.10} &  \secondbest{9.90} & \secondbest{7.33}\\
    \hline
    \end{tabular}
\caption{\textbf{STAGE \textit{vs.} State-of-the-art.} Comparison with state-of-the-art methods on 
     Gaze360 data subsets under the within-dataset setting (Tx = transformer-based TSM). 
    The metric is the mean angular error (in degrees). The \textbf{first} and \underline{second} best 
    results are bold-ed and underlined, respectively.
    }
    \label{table:sota-eval}
\end{table}

\begin{figure*}[t]%
    \centering
    \subfloat[\centering Comparison of Dual-SAM + Tx]{{\includegraphics[width=0.45\textwidth]{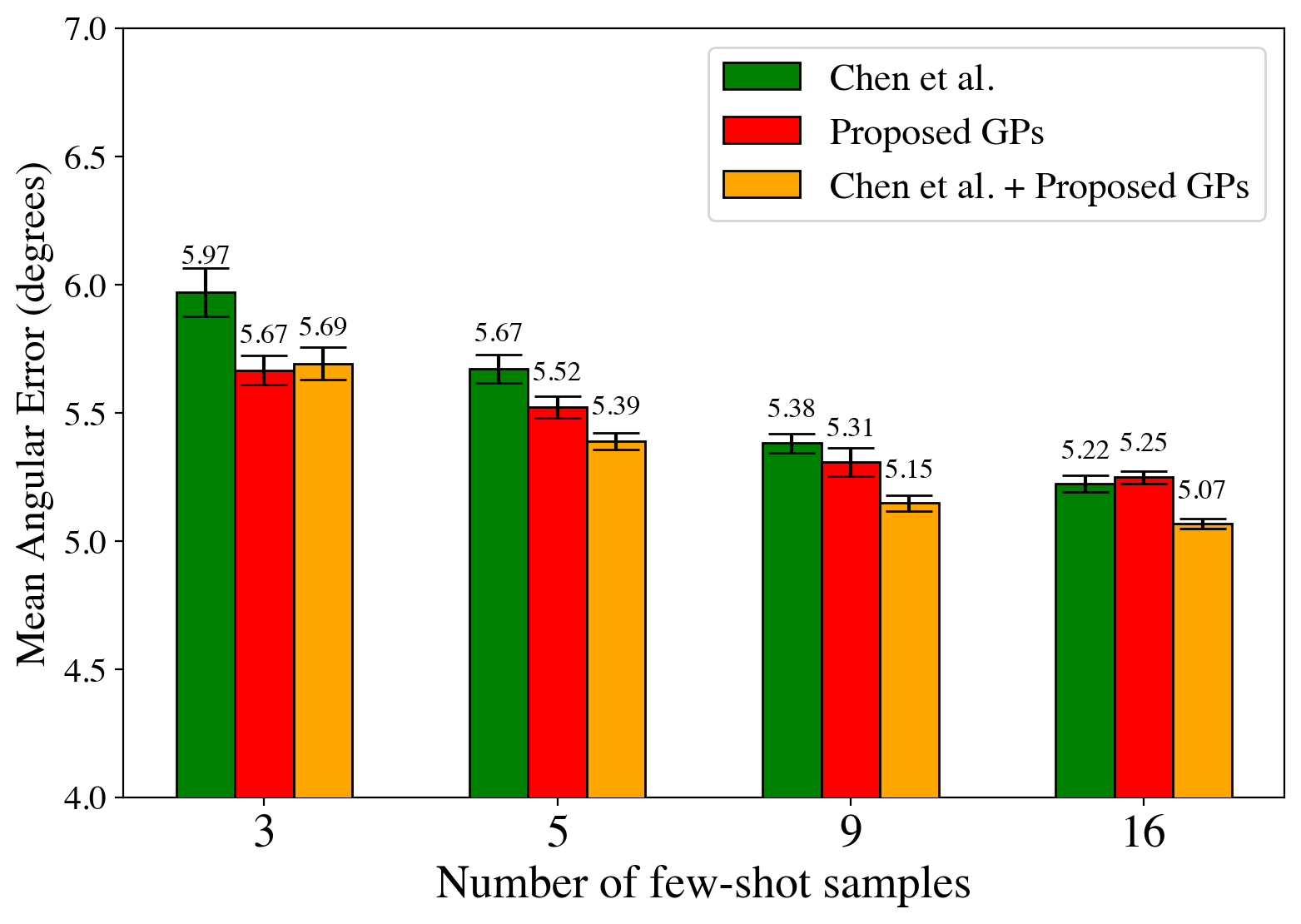} }}%
    \qquad
    \subfloat[\centering Comparison of Hybrid-SAM + Tx]{{\includegraphics[width=0.45\textwidth]{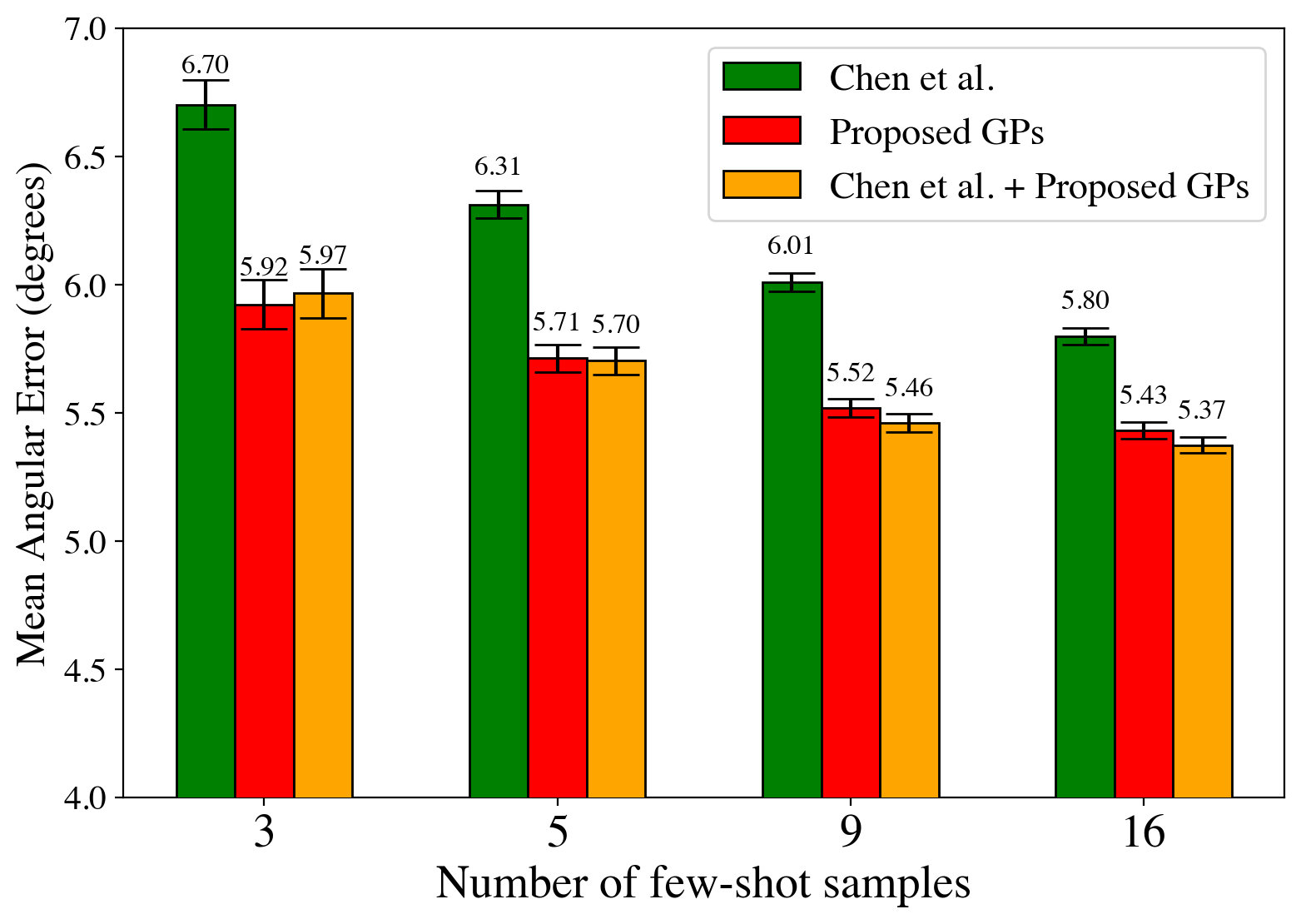} }}%
    \caption{
        The figure shows the comparison of $\ell$-shot GP personalization on the STAGE model with~\citet{chen2020offset} for the EyeDiap dataset. The bars indicate the mean angular error (in degrees) and standard error over 10 iterations. 
        }
    \label{fig:gp-compare}%
\end{figure*}

\subsubsection{Comparison with State-of-the-art Methods}
\label{subsec:sota-exp}
Table~\ref{table:sota-eval} compares the proposed STAGE method with state-of-the-art approaches for a within-dataset setting. Video gaze estimation methods such as the original work of Gaze360~\cite{kellnhofer2019gaze360} and MSA+Seq~\cite{Mishra2020360DegreeGE} employ the LSTM model and learn through the Pinball loss function. We also compare our proposed gaze estimation approach with image-based methods such as L2CS-Net~\citep{abdelrahman2022l2cs}, both variants of GazeTR~\citep{cheng2022gaze}, and self-supervised method SwAT~\citep{farkhondeh2022towards}. We report the performance of these methods from the original work and show a comparison with our method. Our best results outperform these methods by $1.5^{\circ}$, $0.5^{\circ}$ and $2.1^{\circ}$ on full Gaze360, front $180^\circ$ and front $20^\circ$, respectively. 
The superior performance of our method demonstrates the effectiveness of SAM and our choice for other components of the overall STAGE model.

\subsection{Evaluating GPs for Personalization}
\label{sec:gps-eval}

As stated earlier, we first optimize the hyper-parameters of the GP model $\bv{r}_p$ for residual gaze direction prediction using the train subset of EVE dataset. Then, we adapt $\bv{r}_p$ for personalization on the EyeDiap participants. We randomly sample $\ell$ video frames for each participant $10$ times and report the performance in Figure~\ref{fig:gp-compare}. We perform GP personalization on two SAM variants: {Dual-SAM} and {Hybrid-SAM}, using a transformer TSM model.
The baseline method proposed by~\citet{chen2020offset} involves learning a single person-specific bias during training and utilizing a few labeled samples to predict bias during inference.

We obtain an absolute improvement of around $0.8^{\circ}$ with the {Hybrid-SAM} over the baseline with as few as $3$ samples. Applying GPs with the baseline objective, \ie, ``Chen \etal + GPs'', we see consistent improvements over both GPs and the method proposed by~\citet{chen2020offset}. These results demonstrate that GPs' are a valuable tool and provide complementary strengths to~\citet{chen2020offset}. Unlike \citet{chen2020offset}, GPs do not require altering the objective for training the deep network. They can be utilized for adaptation with any pre-trained existing model, such as STAGE.

\begin{figure}[h]%
    \centering
    \subfloat[\centering Examples of certain predictions]{{\includegraphics[width=0.9\columnwidth, height=0.33\columnwidth]{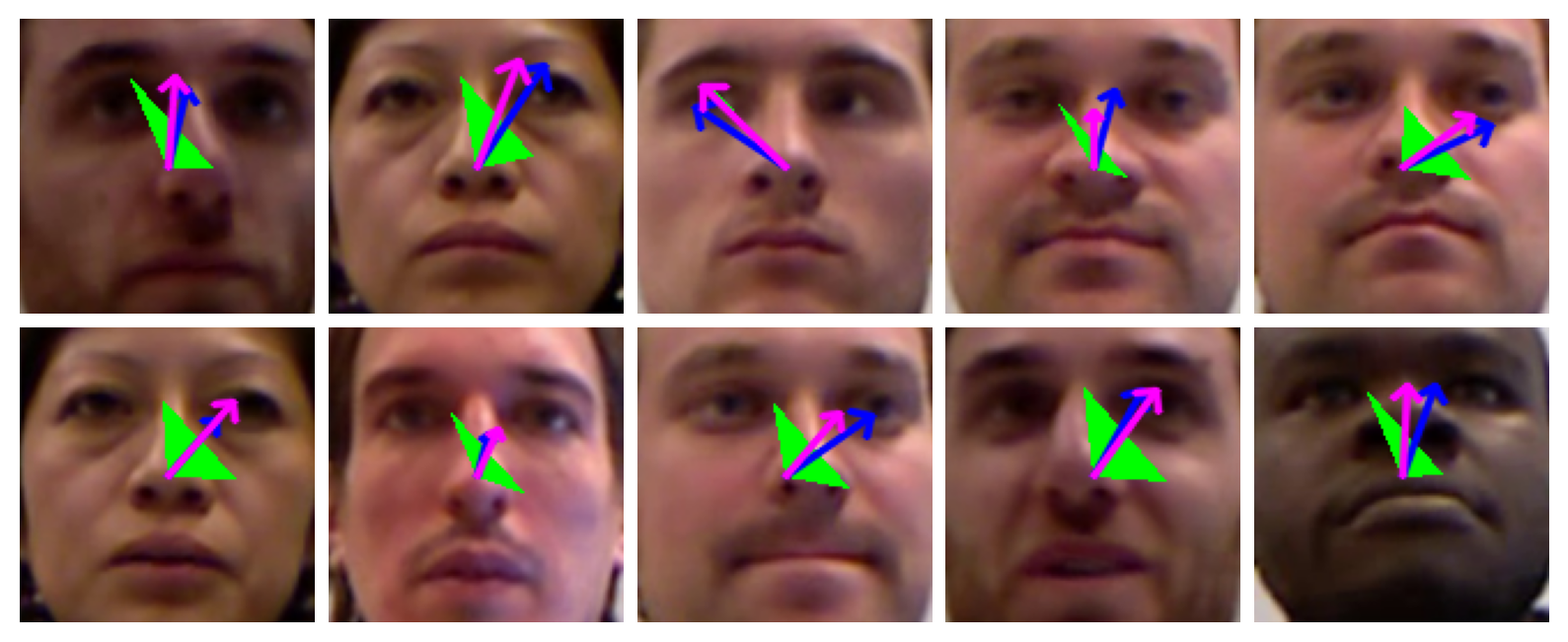} }}%
    \qquad
    \subfloat[\centering Examples of uncertain predictions]{{\includegraphics[width=0.9\columnwidth, height=0.33\columnwidth]{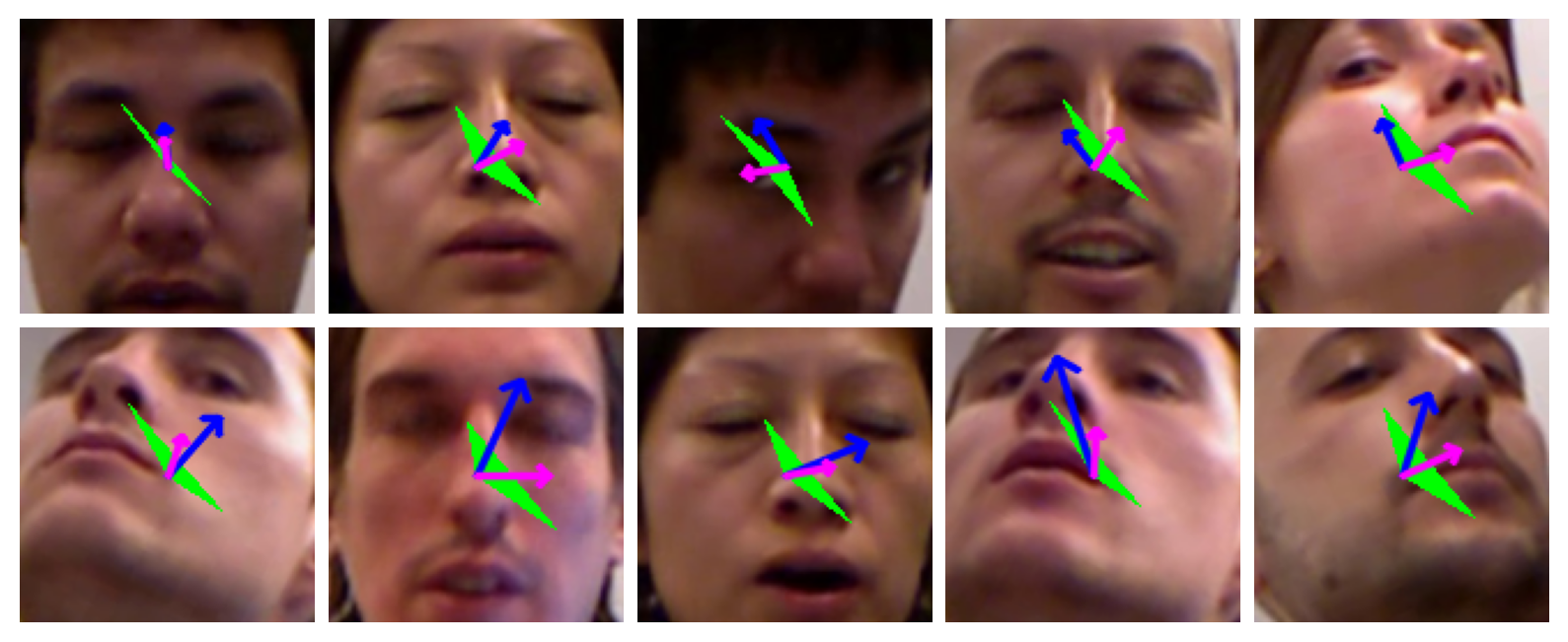} }}%
    \caption{
        The figure depicts a few certain (a) and uncertain (b) predictions for gaze directions after GP's personalization on the EyeDiap dataset. The blue and pink arrows show ground truth and predicted gaze directions, respectively. The green-colored region offers uncertainty of the predictions in the pink arrows. 
        }
    \label{fig:visual}%
\end{figure}

We provide a qualitative evaluation to assess the effectiveness of the GP model's uncertainty, shown in Figure \ref{fig:visual}. The figure shows the differences between confident and uncertain gaze predictions after personalization using the EyeDiap dataset. Notably, the uncertainty region typically includes the ground truth, as illustrated by the pink arrows falling within the green area. It is crucial to note that gaze predictions with higher uncertainty often align with situations that are challenging for human interpretation like extreme head poses or closed eyes.

\section{Conclusion}


In this paper, we presented STAGE, a novel model for video gaze estimation, which utilizes an attention mechanism to encode spatial motion cues and temporal modelling. The method employed a spatial attention module to implicitly focus on the differences between consecutive frames, thereby highlighting relevant changes. We demonstrated that the performance of the STAGE model could be further enhanced using a few labeled samples with Gaussian processes. Future research could explore expanding the receptive field of the attention modules and integrating long-term spatial and temporal dynamics for further enhancements.

{
    \small
    \bibliographystyle{ieeenat_fullname}
    \bibliography{main}
}



\end{document}


\maketitle

\section{Proposed Method
-- Omitted Details}\label{app_sec:proposed_section}

We provide the mathematical formulation of {Dual-SAM} and {Cross-SAM} in Algorithm~\ref{algo:dual-sam} and~\ref{alg:cross_sam}, respectively.

\begin{algorithm}
\caption{Dual-Spatial Attention Module (Dual-SAM)}\label{algo:dual-sam}
\begin{algorithmic}[1]
\REQUIRE  ${X}_{t-1}, {X}_{t} \hfill \in \mathbb{R}^{h \times w \times k}$
\ENSURE  $\mathbf{z}_{t} \hfill \in \mathbb{R}^{3 \cdot k}$ 
\STATE ${X}'_{t-1} = [{X}_{t-1}; {X}_{t}-{X}_{t-1}]$ \newline 
${X}'_{t} \ \ \ = [{X}_{t};{X}_{t}-{X}_{t-1}]  \hfill \in \mathbb{R}^{h\times w \times 2\cdot k}$
\STATE ${A}_{t-1} = \sigma (\conv(\relu(\conv({X}'_{t-1}))))$ \newline 
${A}_{t} = \sigma (\conv(\relu(\conv({X}'_{t})))) \hfill \in \mathbb{R}^{h\times w \times 1} $ 
\STATE $\mathbf{v}_{t-1} = \sum_{h, w} {A}_{t-1} \odot {X}_{t-1}$ \newline 
$\mathbf{v}_{t} = \sum_{h, w} {A}_{t} \odot {X}_{t} \hfill \in \mathbb{R}^{k}$
\STATE $\mathbf{z}_{t} = [\mathbf{v}_{t-1}; \mathbf{v}_{t} - \mathbf{v}_{t-1}; \mathbf{v}_{t}]  \hfill \in  \mathbb{R}^{3 \cdot k}$
\RETURN $\mathbf{z}_{t}$
\end{algorithmic}
\end{algorithm}
\begin{algorithm} 
\caption{Cross-Spatial Attention Module (Cross-SAM)}\label{alg:cross_sam}
\begin{algorithmic}[1]
\REQUIRE  ${X}_{t-1}, {X}_{t} \hfill \in \mathbb{R}^{h\times w\times k}$
\ENSURE  $\mathbf{z}_{t} \hfill \in \mathbb{R}^{3 \cdot d}$
\STATE ${X}_{t-1} =  \flatten( \conv({X}_{t-1}) + \mathbf{1}_{h, w} \odot {P}_{2d})$\newline 
${X}_{t} \ \ \ \ =  \flatten(\conv({X}_{t})  + \mathbf{1}_{h, w} \odot  {P}_{2d}) \hfill \in \mathbb{R}^{h\cdot w\times d}$
\STATE ${X}_{t-1} = \crossattn({X}_{t-1}, {X}_{t}, {X}_{t})$\newline 
${X}_{t} \ \ \ \ = \crossattn({X}_{t}, {X}_{t-1}, {X}_{t-1})  \hfill\in \mathbb{R}^{h\cdot w\times d}$
\STATE $\mathbf{v}_{t-1} = \sum_{h, w} \unflatten({X}_{t-1}, h \times w) $ \newline 
$\mathbf{v}_{t} = \sum_{h, w} \unflatten({X}_{t}, h \times w)\hfill\in \mathbb{R}^{d}$
\STATE $\mathbf{z}_{t} = [\mathbf{v}_{t-1}; \mathbf{v}_{t} - \mathbf{v}_{t-1}; \mathbf{v}_{t}]  \hfill \in  \mathbb{R}^{3 \cdot d}$
\RETURN $\mathbf{z}_{t}$
\end{algorithmic}
\end{algorithm}

\begin{figure}[h]
    \centering
    \includegraphics[scale=0.4]{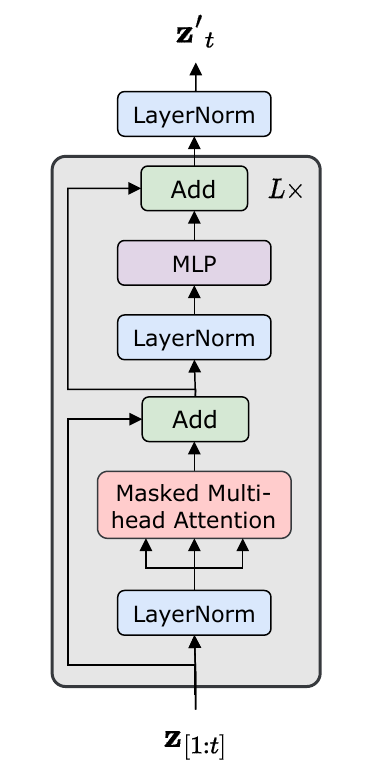}
    \caption{Block diagram of transformer temporal sequence model.} 
    \label{fig:blocktsm}
\end{figure}

\paragraph{Transformer Block.} Figure~\ref{fig:blocktsm} shows the architecture of a single transformer layer used in the temporal sequence model of the STAGE method. MLP is a Multi-Perceptron layer, and we use $L$ layers stacked together in the TSM.

We incorporate learned temporal position embeddings to enable the transformer model to discern temporal relationships within the input feature sequence. These embeddings are uniquely associated with each position, providing the model with explicit information about the relative ordering of elements within the sequence.
The embedded features are then passed through multiple layers, each consisting of masked multi-head attention, LayerNorm (LN), and MLP. Masked multi-head attention allows the transformer model to attend to only past frame features. The output of the TSM is a feature sequence passed through an LN layer, similar to the GPT-2 model~\citep{radford2019language}.

\section{Additional Implementation Details}
\label{app:imple}
The {Dual-SAM} consists of two convolutional layers with kernel size $1$ and output feature maps of $64$ and $1$, respectively. The first convolutional layer has a group normalization layer~\citep{wu2018group} applied to the output features, followed by a dropout layer with $p=0.5$. 
In {Cross-SAM} and {Hybrid-SAM}, we project the incoming features to higher channels through a convolution layer with $d=512$ and a kernel size $1$.  After adding 2D positional embeddings to the projected feature maps, they go through the cross-attention encoder, which consists of four heads and two layers with an embedding size of $64$.

The TSM model has two variants: an LSTM variant and a transformer variant. The LSTM variant consists of one unidirectional LSTM layer with a hidden dimension of $128$.  The transformer variant is based on GPT-2~\citep{radford2019language} network with $6$-heads and $6$-layers, operating on a dimension of $d=128$, and initialized randomly. The gaze prediction layer consists of two fully connected (FC) layers. The first FC layer has a SeLU activation function and a hidden dimension of the same size as the input dimension. The second FC layer outputs the 2D gaze direction angles, pitch and yaw.

Our STAGE model is implemented in PyTorch~\citep{paszke2017automatic}. We set $\lambda=0.001$ for cross-data and $\lambda=0$ for within-data evaluations. For GP hyper-parameter optimization, we use Adam optimizer with a learning rate of $0.001$, implemented using GPytorch~\citep{gpytorch}. Our code and trained models will be made publicly available in the future and are zipped in supplementary.

\section{Ablation Study}
\label{sec:ablation}
In the ablation study, we study the impact of adding multiple SAM blocks in the STAGE model, where the output of one SAM goes as input to the next. The ablation study on the number of Dual- and Hybrid-SAM blocks (four blocks vs. one block) for within-data and cross-data settings are shown in Tables~\ref{table:within-multisam}(a) and (b), respectively. We observe no significant improvements over a single block of SAM, indicating that one SAM block is enough to provide spatial motion cues between consecutive frame features and improve performance.

\begin{table}[t]
\centering
\centering
\subfloat[Within-dataset evaluation]{
\small
\centering
\resizebox{\columnwidth}{!}{%
\begin{tabular}{l|c|c|c} 
\hline
Method & Full & $180^\circ$ & $20^\circ$\\
\hline

Dual-SAM(1-block)+Tx & 10.13 & 9.93 & 7.23\\

Hybrid-SAM(1-block)+Tx & 10.10 &  9.90 & 7.33\\
\hline

Dual-SAM(4-blocks)+Tx  & 12.13 & 11.68 & 9.33\\
Hybrid-SAM(4-blocks)+Tx & 10.25 & 10.08 & 7.27 \\
\hline  
\end{tabular}
}
}
\\
\vspace{5pt}
\subfloat[Cross-dataset evaluation]{
\small
\centering
\resizebox{\columnwidth}{!}{%
\begin{tabular}{l|c|c|c} 
\hline
Method & EyeDiap & Full & $180^\circ$ \\ 
\hline

Dual-SAM(1-blocks)+Tx & 6.77 & 23.99 & 23.38 \\ 

Hybrid-SAM(1-blocks)+Tx & 6.54 & 23.77 & 23.17  \\ 

\hline

Dual-SAM(4-blocks)+Tx   & 7.27 & 23.34 & 22.74  \\ 
Hybrid-SAM(4-blocks)+Tx & 7.55 &  23.52 & 22.91 \\ 
\hline  
\end{tabular}
}
}
\caption{\textbf{Ablation Study:} Comparison of different numbers of SAM blocks employed in our STAGE method. Tx is transformer-based TSM, and training is performed for within-data and cross-data settings in (a) and (b), respectively. The metric reported is mean angular errors (in degrees).}
\label{table:within-multisam}
\end{table}

\section{Additional Results for GP Evaluation}

\begin{figure} 
  \centering
  \includegraphics[scale=0.4]{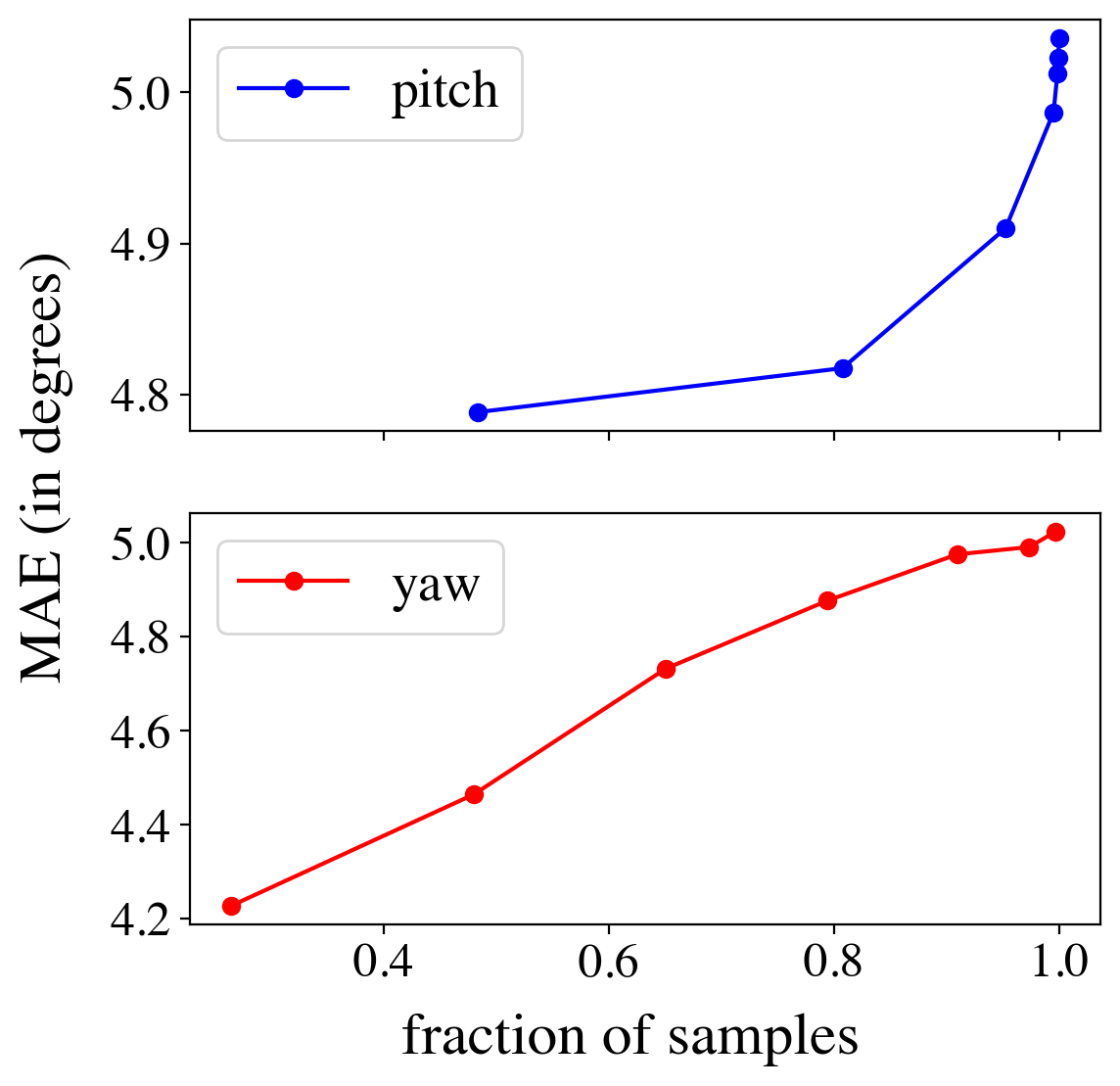}
  \caption{Comparison of Mean Angular Error (in degrees) of gaze components (yaw or pitch) with increasing fraction of test samples sorted with respect to the uncertainty of GP predictions. Plots exhibit that GPs are more accurate when the prediction is relatively more confident (with less variance).}
  \label{fig:percimg}
\end{figure}

For assessing the effectiveness of the GP model's uncertainty, we provide additional analysis of gaze predictions, as illustrated in Figure \ref{fig:percimg}. Our evaluation begins with an analysis of the GP's posterior variance diagonal. We arrange this in ascending order and then apply different uncertainty thresholds to it.
For each selected threshold, we compute the MAE on test samples that exhibit uncertainty levels below the threshold. This procedure is repeated across a range of different thresholds to evaluate performance. Figure \ref{fig:percimg} presents a comparison of the MAE for yaw and pitch against increasing fractions of test data samples. These samples are sorted according to the uncertainty in the GP prediction. This analysis demonstrates that GPs tend to deliver more accurate results when their variance is lower, signifying greater confidence in the predictions. Therefore, the uncertainty measure in the GP model can act as an effective indicator to avoid making inaccurate predictions.

\begin{figure*}[!tbp]
\captionsetup[subfigure]{aboveskip=-40pt}
  \begin{subfigure}[b]{\linewidth}
    \includegraphics[width=\linewidth]{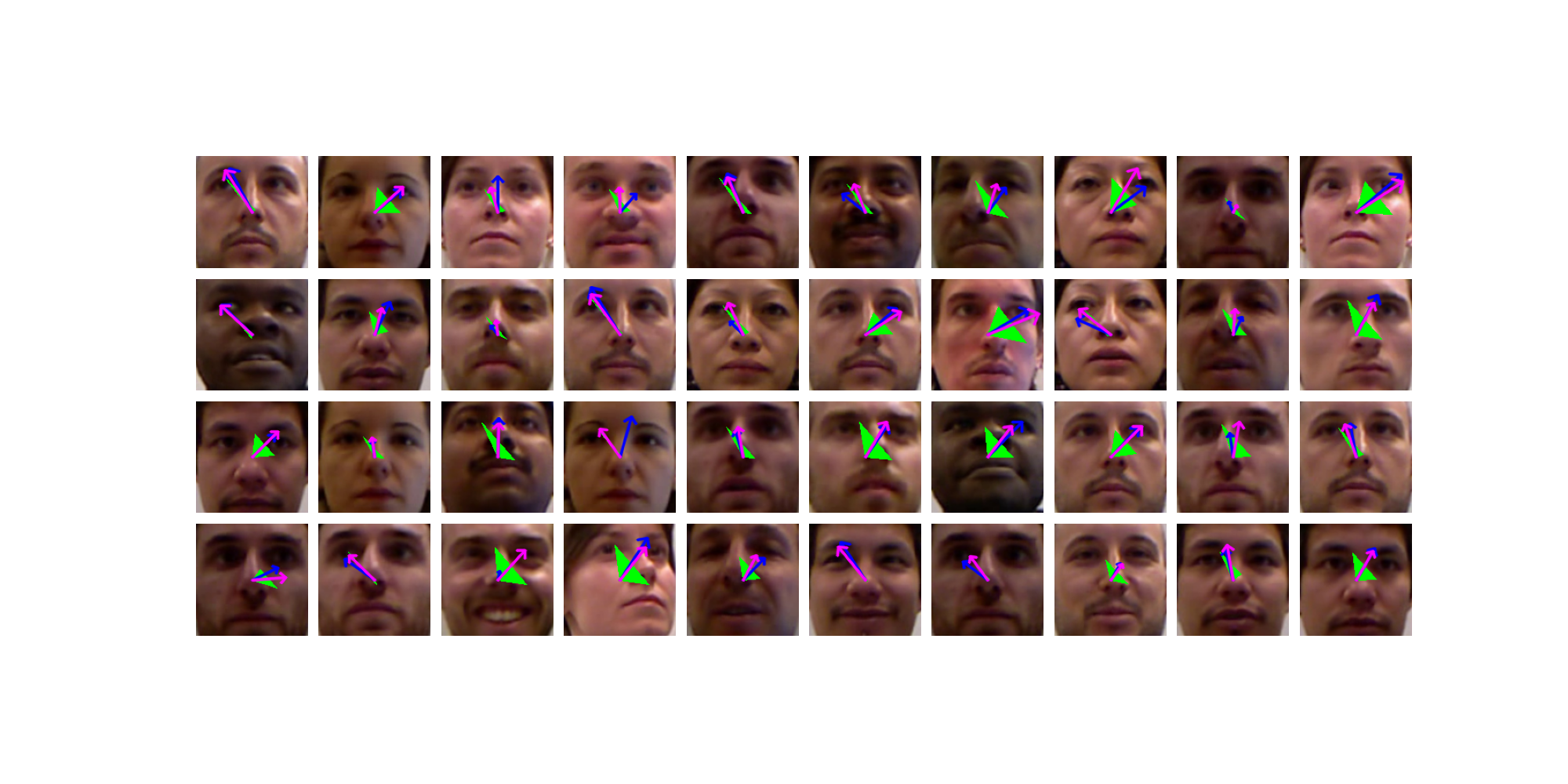}
    \caption{Certain predictions for EYEDIAP dataset}
    \label{fig:f1}
  \end{subfigure}
  \hfill
  \begin{subfigure}[b]{\linewidth}
    \includegraphics[width=\linewidth]{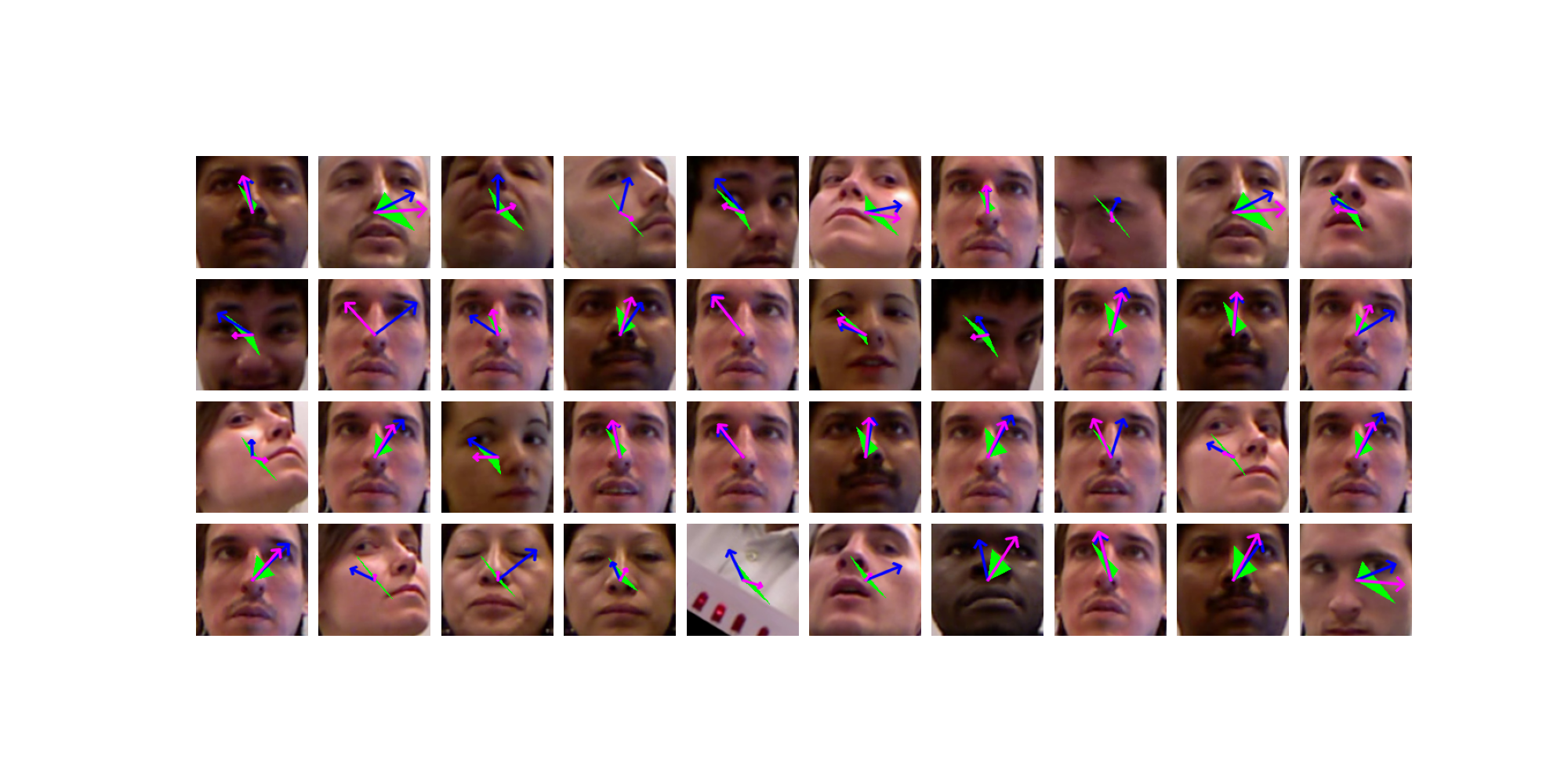}
    \caption{Uncertain Predictions for EYEDIAP dataset}
    \label{fig:f2}
  \end{subfigure}
  \caption{The figure depicts a few confident \ref{fig:f1} and uncertain \ref{fig:f2} predictions for gaze directions after GP's personalization on the EYEDIAP dataset. Blue and pink arrows show ground truth and predicted gaze directions, respectively. The green-colored region offers uncertainty of the predictions in pink arrows. The uncertainty region often covers the ground truth, \ie, the pink arrows are in the green-colored area.}
\end{figure*}

We also provide additional visualizations of the predictions from personalized GP on top of the STAGE model, similar to Figure 6 in the main manuscript. Figure~\ref{fig:f1} and \ref{fig:f2} respectively show certain and uncertain prediction images from the EYEDIAP dataset after performing GP personalization. The ground truth and predicted gaze directions are respectively shown with blue and pink colored arrows, and the corresponding uncertainty of prediction is shown with the green colored triangle.

{
    \small
    \bibliographystyle{ieeenat_fullname}
    \bibliography{main}
}
